\documentclass[conference]{IEEEtran} 
\usepackage{cite}
\usepackage{amsmath}
\usepackage{amsfonts}
\usepackage{setspace}
\usepackage{amsthm}
\usepackage{tabularx}
\usepackage{graphicx}
\usepackage{enumerate}
\usepackage{multirow}
\usepackage{empheq}
\usepackage{textcase}
\usepackage{algorithmic}
\usepackage{textcomp}
\usepackage{xcolor}
\usepackage{amsmath,amssymb,amsfonts}
\usepackage{url}
\usepackage{flushend}
\usepackage{subcaption}
\usepackage{hyperref}

\IEEEoverridecommandlockouts

\DeclareRobustCommand{\IEEEauthorrefmark}[1]{\smash{\textsuperscript{\footnotesize #1}}}

\begin{document}

\title{AV4EV: Open-Source Modular Autonomous\\Electric Vehicle Platform\\for Making Mobility Research Accessible}


\makeatletter 
\newcommand{\linebreakand}{%
  \end{@IEEEauthorhalign}
  \hfill\mbox{}\par
  \mbox{}\hfill\begin{@IEEEauthorhalign}
}
\makeatother 

\author{
    \IEEEauthorblockN{
        Zhijie Qiao\IEEEauthorrefmark{1,2}\textsuperscript{*},
        Mingyan Zhou\IEEEauthorrefmark{1}\textsuperscript{*},
        Zhijun Zhuang\IEEEauthorrefmark{1},
        Tejas Agarwal\IEEEauthorrefmark{1,2},
        Felix Jahncke\IEEEauthorrefmark{1,3}, \\
        Po-Jen Wang\IEEEauthorrefmark{2},
        Jason Friedman\IEEEauthorrefmark{1,2}, 
        Hongyi Lai\IEEEauthorrefmark{1,2},
        Divyanshu Sahu\IEEEauthorrefmark{1},
        Tomáš Nagy\IEEEauthorrefmark{1,4},
        Martin Endler\IEEEauthorrefmark{1,4}, \\
        Jason Schlessman\IEEEauthorrefmark{2,5}, 
        Rahul Mangharam\IEEEauthorrefmark{1,2}
    }

    \IEEEauthorblockA{
        \IEEEauthorrefmark{1}School of Engineering and Applied Science, University of Pennsylvania, Email: rahulm@seas.upenn.edu \\ 
        \IEEEauthorrefmark{2}Autoware Foundation Center of Excellence for Autonomous Driving \\
        \IEEEauthorrefmark{3}Professorship of Autonomous Vehicle Systems, Technical University of Munich\\
        \IEEEauthorrefmark{4}Czech Institute of Informatics, Robotics and Cybernetics, Czech Technical University in Prague \\
        \IEEEauthorrefmark{5}Red Hat Research \\
    }
}

\maketitle
\begingroup\renewcommand\thefootnote{*}
\footnotetext{These authors contributed equally to this work.}
\endgroup

\begin{abstract}
When academic researchers develop and validate autonomous driving algorithms, there is a challenge in balancing high-performance capabilities with the cost and complexity of the vehicle platform. Much of today's research on autonomous vehicles (AV) is limited to experimentation on expensive commercial vehicles that require large skilled teams to retrofit the vehicles and test them in dedicated facilities. 
On the other hand, 1/10th-1/16th scaled-down vehicle platforms are more affordable but have limited similitude in performance and drivability. 
To address this issue, we present the design of a one-third-scale autonomous electric go-kart platform with open-source mechatronics design along with fully functional autonomous driving software.
The platform's multi-modal driving system is capable of manual, autonomous, and teleoperation driving modes.
It also features a flexible sensing suite for the algorithm deployment across perception, localization, planning, and control. 
This development serves as a bridge between full-scale vehicles and reduced-scale cars while accelerating cost-effective algorithmic advancements. 
Our experimental results demonstrate the AV4EV platform's capabilities and ease of use for developing new AV algorithms. 
All materials are available at AV4EV.org to stimulate collaborative efforts within the AV and electric vehicle (EV) communities.
\end{abstract}



\section{INTRODUCTION}
\label{sec:intro}

The increasing interest in self-driving cars has ushered in a new area of study in recent years: autonomous racing. This involves the development of software and hardware for high-performance racing vehicles intended to function autonomously at unprecedented levels, including high speeds, substantial accelerations, minimal response times, and within unpredictable, dynamic, and competitive settings \cite{JohannesMotorsport}. 
However, a significant hurdle remains the unavailability of full-sized vehicles and the accessibility of smaller-scaled RC cars. 
For full-sized platforms that encompass independent driving capacities such as the Dallara AV21 from Indy Autonomous Challenge \cite{IAC}, testing the limits of safety and performance is costly and hazardous, and also outside the reach of most academic departments and research groups. 
For smaller-scaled RC cars such as F1TENTH \cite{o2020f1tenth}, the limited capability of sensing and computing constrains the complexity of the algorithms and the level of research conducted.

To address this issue, we created AV4EV, an accessible, open-source reference model for a one-third-scale autonomous electric racing platform. This platform merges the capabilities of full-sized vehicles with the compactness and adaptability of its smaller size. AV4EV offers open-source designs for mechatronics, sensing, and autonomous driving software, aiming to provide a standardized solution for modular autonomous and electric vehicles.

Our go-kart won the championship at the 2023 Autonomous Karting Series Purdue Grand Prix, where it competed against several other US national teams \cite{AKS}.
This autonomous go-kart solution can easily be adopted by universities and research institutes to promote the safe and effective development and verification of AV.

\begin{figure*}[t]
    \centering
    \includegraphics[width=\textwidth]{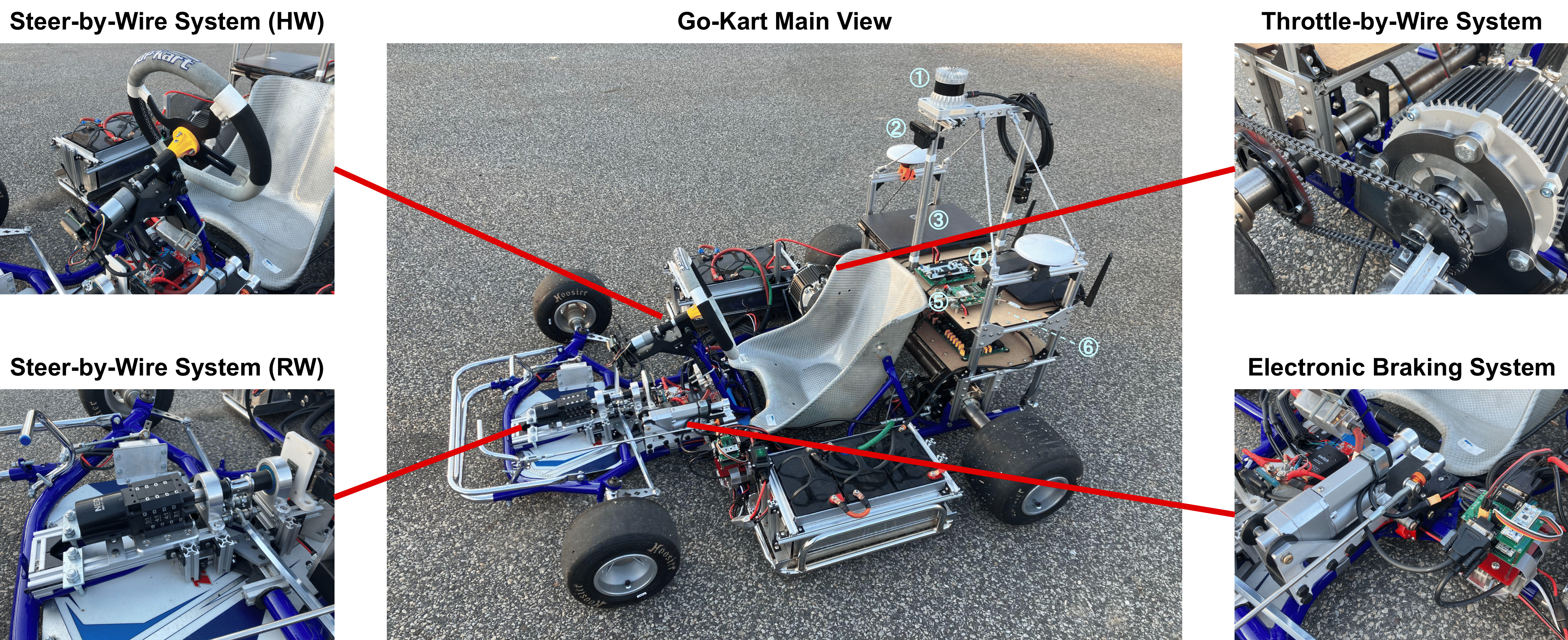}
    \caption{Go-kart platform overview with Steer-by-Wire System (SBWS) including its hand wheel (HW) and road wheel (RW) components, Throttle-by-Wire System (TBWS), and Electronic Braking System (EBS). The sensors and computing units mounted on the double-deck rear shelf are enumerated from top to bottom as follows: (1) Ouster LiDAR, (2) OAK-D camera, (3) Onboard laptop, (4) Main Control System (MCS), (5) Sepentrio GNSS, and (6) IMU, concealed from the main view perspective, is positioned on the lower deck.}
    \label{fig:control_modes}
    \vspace{-0.5cm}
\end{figure*}

This work makes the following contributions:
\begin{enumerate}
    \item We introduced an accessible modular electric vehicle platform with multi-driving modes (manual, autonomous, and teleoperated), bridging the gap between full-scale vehicles and RC cars. The estimated cost of constructing one go-kart, including all mechatronic systems, stands at approximately 12,500 USD. It is expected that with scaled production, the cost will decrease substantially.
    \item We developed a flexible sensing suite and demonstrative software solutions to handle autonomous driving capabilities validated through experiments. The estimated cost is around 11,000 USD, while the figure can vary depending on user-specific requirements and customization.
    \item We provided comprehensive open-source resources to guide building and testing the one-third-scale electric go-kart with detailed tutorials, GitHub repositories for hardware design and software stacks, demonstration videos, a bill of materials \cite{gokart_doc, gokart_mechatronics_git, gokart_sensor_git}.
\end{enumerate}

\section{Mechatronics}

The go-kart mechatronic system is designed as a modular system, consisting of several subsystems that are responsible for different vehicle execution tasks. There are five subsystems which integrated with the base go-kart chassis in a non-intrusive way: Power Distribution System (PDS), Main Control System (MCS), Throttle-by-Wire System (TBWS), Steer-by-Wire-System (SBWS), and Electronic Braking System (EBS) (Fig. \ref{fig:control_modes}).
All subsystems except the PDS utilize an STM32 Nucleo development board on a standalone PCB as the electronic control unit (ECU). Communication among these modular systems is achieved through the controller area network (CAN), aligning with modern vehicle design standards for efficient information exchange.

\subsection{Power Distribution System (PDS)}
The autonomous go-kart is powered by six Nermak Lithium LiFePO4 deep cycle batteries, each possessing a voltage of 12V and a capacity of 50Ah. These batteries are installed on both sides of the go-kart and interconnected via wiring across the chassis. Four of them are linked in a series, yielding a net voltage of 48V, which powers the TBWS motor. A step-down converter is utilized to convert the voltage from 48V to 12V, which in turn provides power to the SBWS and EBS motor. The remaining two batteries, also interconnected in series, produce a net voltage of 24V. This voltage is then fed through several converters to obtain different desired voltages to power up the sensing (Fig. \ref{fig:pdu_sub1}) and control (Fig. \ref{fig:pdu_sub2}) systems.  

\subsection{Main Control System (MCS)}
The MCS handles all driving requests from the top-level supervisory controller and dispatches commands (throttle, steering, brake) on the CAN bus \cite{CANbus}. It serves as an interface between the go-kart mechatronic system and the end user. Three different operation modes are supported: manual, remote, and autonomous. In manual mode, input is read from the steering wheel, throttle, and brake pedals of a driver, just like in a conventional vehicle. In remote mode, the operator uses a Spektrum DX6 2.4GHz radio to send driving commands to the MCS. In autonomous mode, the command is transmitted from a high-level computing unit, such as a laptop or an onboard computer, through USB-to-TTL communication. After receiving the desired driving commands, the MCS sends these on the CAN bus to be received by the subsystems. Meanwhile, each subsystem measures its state with sensors and sends feedback on the CAN bus. This feedback is gathered by the MCS and shared with the operator.

\subsection{Throttle-by-Wire System (TBWS)}
The TBWS includes the electronic controller unit (ECU) and VESC 75/300 motor driver to control the go-kart's main drive motor. The brushless DC motor (ME1717 from Motenergy) transmits the motion to the go-kart rear axle of through a chain and drives the wheels in the longitudinal direction. The ECU receives the desired speed from the MCS via the CAN bus, measures the current speed through an encoder, and outputs the desired throttle signal to the VESC controller, which then powers up the motor. Additionally, a remote kill switch is added independent of the ECU that allows the user to kill power, thus ensuring safety in the worst-case scenario. 

\begin{figure*}[t!] 
    \centering
    \begin{subfigure}{.5\textwidth}
        \centering
        \includegraphics[width=.8\linewidth]{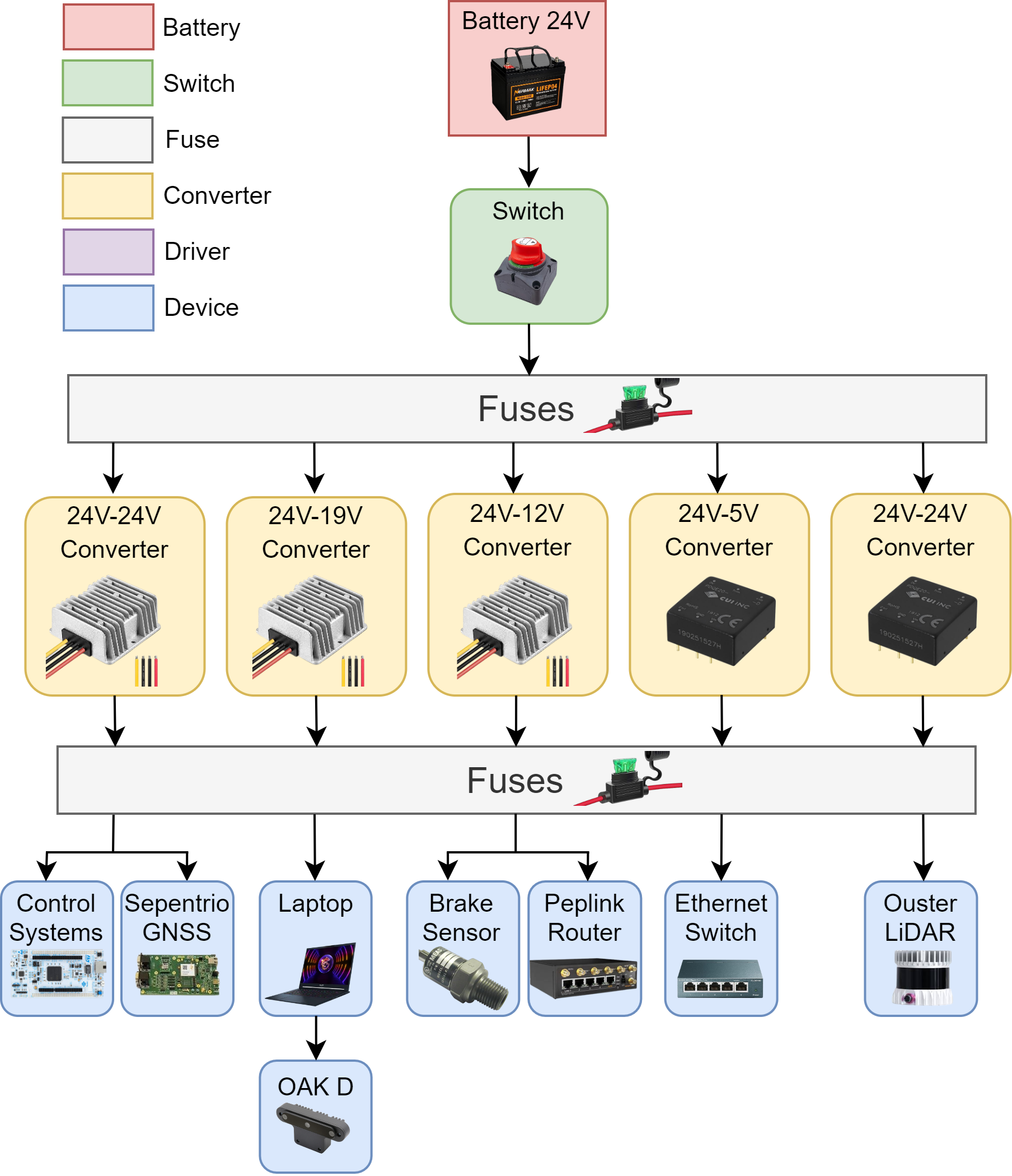}
        \caption{Sensing}
        \label{fig:pdu_sub1}
    \end{subfigure}%
    \begin{subfigure}{.5\textwidth}
        \centering
        \includegraphics[width=.8\linewidth]{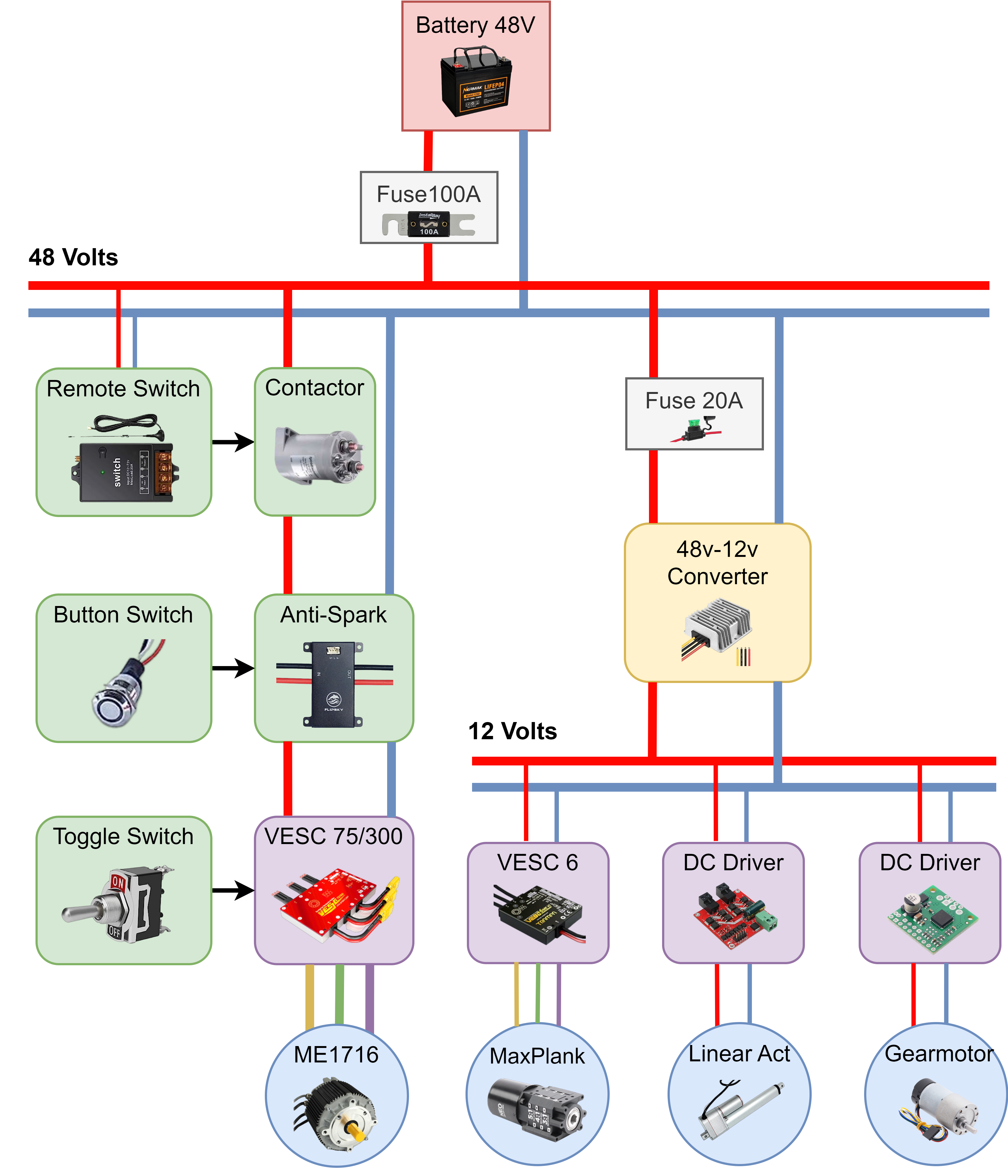}
        \caption{Motor}
        \label{fig:pdu_sub2}
    \end{subfigure}
    \caption{Sensing (left) and motor (right) power system with connections and devices. }
    \label{fig:whole}
    \vspace{-0.5cm}
\end{figure*}

\subsection{Steer-by-Wire System (SBWS)}
The SBWS eliminates the mechanical steering shaft between the hand wheel (HW) and road wheel (RW), allowing each part to be governed by its motor, sensor, and ECU \cite{kaufmann2001SBWdevelopment}. This design reduces weight, space, and cost with the modular structure, while improving the flexibility and availability of autonomous driving functions \cite{SBWdev}. 
Our HW component utilizes a brushed DC motor to coaxially drive the HW. 
The RW component employs a NEO1650 Brushless DC motor to propel the two front wheels via steering tie rods as linkages. 

\subsection{Electronic Braking System}
The original go-kart design translates movement from the driver pressing the brake pedal to the master cylinder and reservoir via the push rod, generating hydraulic braking pressure without the need for additional servo motors.
To achieve autonomous braking without human input, a linear actuator is mounted at the end of the push rod to create a linear movement simulating the pedal-pressing action. 
This non-intrusive design allows the safety operator (if present) to press the brake pedal regardless of the linear actuator state.
Finally, a pressure sensor is installed onto the braking hydraulic system to collect data for effective feedback control.


\section{Sensing}

The sensing system is a fundamental module for research and development for perception and localization. Our design employs a flexible sensor setup that can be customized and reconfigured to suit different objectives and priorities.

To start up, an Ouster OS1 LiDAR is positioned at the highest point on the rear end of the go-kart to leverage its max 200-meter range and 360-degree field of view. The OAK-D camera, placed below the LiDAR, has the capabilities of high-resolution image capturing, depth measuring, and long-range object tracking. These features work seamlessly with the LiDAR point cloud for object fusion and post-processing.  

Moreover, the go-kart is equipped with a Global Navigation Satellite System (GNSS) and an Inertial Measurement Unit (IMU). 
For GNSS, we utilized the Sepentrio Mosaic-H carrier board with two Multiband antennas (IP66) from ArduSimple mounted on both rear sides of the go-kart. We also subscribed to Swift Navigation’s real-time kinematic positioning (RTK) service, enabling our GNSS to achieve centimeter-level position accuracy. 
In situations where GNSS signals are disrupted due to severe weather or signal obstructions, an IMU is needed for localization filtering. Thus, we placed a BNO055 9-DOF IMU on the go-kart's centerline of mass to provide accurate accelerometer, gyroscope, and magnetometer information. 

All sensors transmit data to an onboard laptop, which then executes algorithms and transmits drive commands to the MCS. We used the MSI Pulse GL66 15.6" Gaming Laptop, integrated with an Intel Core i7-12700H, an RTX3070 GPU, 16GB of internal RAM, and a storage capacity of 512GB. The laptop also contains three USB 3.0 ports and one Ethernet port to support high-speed data transmission with the sensors.

\section{Software}

\begin{figure*}[t]
    \centering
    \includegraphics[width=\textwidth]{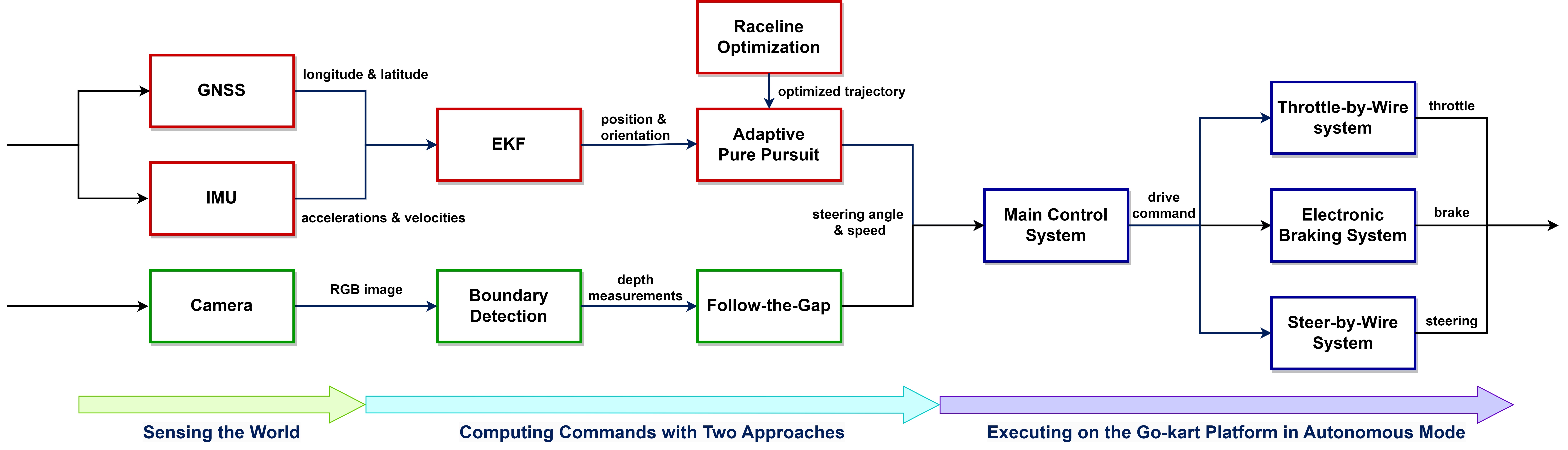}
    \caption{Software pipeline for go-kart autonomous driving capabilities: GNSS-based adaptive pure pursuit (red), camera-based follow-the-gap (green), go-kart mechatronics execution (blue).}
    \label{fig:pipeline}
    \vspace{-0.5cm}
\end{figure*}

We designed an autonomous racing framework using the Robot Operating System (ROS2) for our go-kart platform with the free tooling of Python and C++. This framework incorporates two primary algorithms: a GNSS-based pure pursuit method for pre-mapped racing and a camera-based follow-the-gap (FTG) algorithm \cite{SEZER2012FGM} for reactive racing.
In the framework outlined in this paper, LiDAR was not used due to its complexity. Nevertheless, the LiDAR data is readily available and will be integrated into future research. The software pipeline within the holistic autonomous driving workflow is illustrated in Fig. \ref{fig:pipeline}. 


\subsection{Localization}

The position measurements from the GNSS are presented in latitude and longitude. To convert these geographical coordinates into a more interpretable format within a local frame, we utilized the equirectangular projection method as in equations \eqref{eq:equi_xy}:
\begin{equation}
x = r \cdot \cos(lat) \cdot lon, \ y = r \cdot lat,
\label{eq:equi_xy} 
\end{equation}
where \(r\) symbolizes the mean radius of the Earth, which is 6371 kilometers, \(lat\) stands for latitude (radians), and \(lon\) denotes longitude (radians). A reference point is first established, and all subsequent coordinates are defined with respect to this reference point, treating it as the origin \cite{gnss_to_local}. 
While this approach has the potential to introduce distortion, in our case, the impact is negligible due to the small size of the testing field.

As previously mentioned, there are instances where the GNSS signal may experience interruptions. 
To guarantee timely and accurate localization information, we implemented an Extended Kalman Filter (EKF) that integrates IMU data.
Evolving dynamically over time \(t\), the velocity motion model \(X_t\) adopted for the go-kart is consisting of the position \(x_t, y_t\) and orientation \(\psi_t\); the input to the system is linear velocity \(v_t\) and angular velocity \(\omega_t\); and the identity covariance matrix \(P_t\) that is initialized at timestamp zero:
\begin{equation}
X_t = [x_t, y_t, \psi_t]^T,
\end{equation}
\begin{equation}
u_t = [v_t, \omega_t],
\end{equation}
\begin{equation}
P_t = \begin{bmatrix}
\sigma_x^2 & \sigma_{xy} & \sigma_{x\psi} \\
\sigma_{yx} & \sigma_y^2 & \sigma_{y\psi} \\
\sigma_{\psi x} & \sigma_{\psi y} & \sigma_\psi^2 \\
\end{bmatrix} .
\end{equation}
At timestamp \(t\), the system is linearized around the current state, and the prediction step is executed as follows:
\begin{equation}
X_{t+1|t} = 
\begin{bmatrix} 
x_t + v_t \Delta t \cdot cos(\psi_t)\\
y_t + v_t \Delta t \cdot sin(\psi_t)\\
\psi_t + \Delta t \omega_t\\
\end{bmatrix} .
\end{equation}
For each state in the system, we calculated the partial derivatives with respect to the other states to obtain the Jacobian matrix:
\begin{equation}
J = \begin{bmatrix} 
1 & 0 & -\Delta t \cdot v \cdot \sin(\psi) \\
0 & 1 & \Delta t \cdot v \cdot \cos(\psi) \\
0 & 0 & 1 \\
\end{bmatrix} .
\end{equation}
The prediction update of the covariance matrix is as follows: 
\begin{equation}
P_{t+1|t} = J P J^T + R,
\end{equation}
where the dynamic noise \(R\) is approximated as a constant diagonal matrix of \(0.1\), with units in meters and radians.

For the observation step, we extracted position data \(x\) and \(y\) from the GNSS and orientation data \(\psi\) from the IMU, and denote them with subscripts:
\begin{equation}
X_{obs} = [x_{obs}, y_{obs}, \psi_{obs}]^T.
\end{equation}
Given that the observation directly corresponds to the state, the Jacobian is equivalent to the identity matrix. By combining the variance readings from the sensors that are organized as a diagonal matrix \(M\), we could calculate the Kalman gain \(K\):
\begin{equation}
M = \begin{bmatrix}
\sigma_{xobs}^2 & 0 & 0 \\
0 & \sigma_{yobs}^2 & 0 \\
0 & 0 & \sigma_{\psi obs}^2 \\
\end{bmatrix},
\end{equation}
\begin{equation}
K = P_{t+1|t} I^T (IP_{t+1|t}I^T+M)^{-1},
\end{equation}
Finalize the update step to complete localization:
\begin{equation}
X_{t+1|t+1} = X_{t+1|t} + K(X_{t+1|t} - X_{obs}),
\end{equation}
\begin{equation}
P_{t+1|t+1} = (I - KI)P_{t+1|t}.
\end{equation}

\subsection{Raceline Optimization}

In pre-mapped racing scenarios, a reference racing line is generally acquired in advance and subsequently tracked by the controller. The raceline is represented by a sequence of waypoints consisting of the target position \(x\), \(y\), velocity \(v\), etc. 
While manually piloting the go-kart, waypoints are gathered at consistent temporal or spatial intervals. It may not account for vehicle dynamics, resulting in a non-smooth trajectory.

Therefore, we integrated a min-curvature raceline optimization algorithm as proposed in \cite{MINCURVE}. 
First, we calibrated the physical properties of the go-kart such as mass, width, maximum turning radius, maximum acceleration, etc., assuming a track of uniform width, and then utilized manually collected waypoints to depict the centerline of the track. The raceline points can be parameterized as:
\begin{equation}
\vec{r}_i = \vec{p}_i + \alpha_i \vec{n}_i,
\end{equation}
where \(\vec{p}_i = [ x_i, y_i ]^T\) is the center line point, \(\vec{n}_i\) is the unit length normal vector, and \(a_i\) encodes the track boundaries. The raceline is then defined through third-order spline interpolations of the points \(r_i\) in \(x\) and \(y\) coordinates. Followed the formulation in \cite{MINCURVE}, we minimized the discrete squared curvature \(\gamma_i\) of the splines along the raceline:
\begin{align}
& \underset{[ \alpha_1 \cdots \alpha_N ]}{\text{minimize}} & & \sum_{i=1}^{N} \gamma_i^2(t) \\
& \text{subject to} & & \alpha_i \in [\alpha_{i,\text{min}}, \alpha_{i,\text{max}}] & & \forall 1 \leq i \leq N.
\end{align}

Subsequently, we generated the velocity profile considering the longitudinal and lateral acceleration limits of the car at various velocities. As depicted in Fig. \ref{fig:raceline}, the optimized raceline shows reduced curvature, thereby enhancing smoothness and eliminating overlapping waypoints.

\begin{figure}[!t]
    \includegraphics[width=\columnwidth]{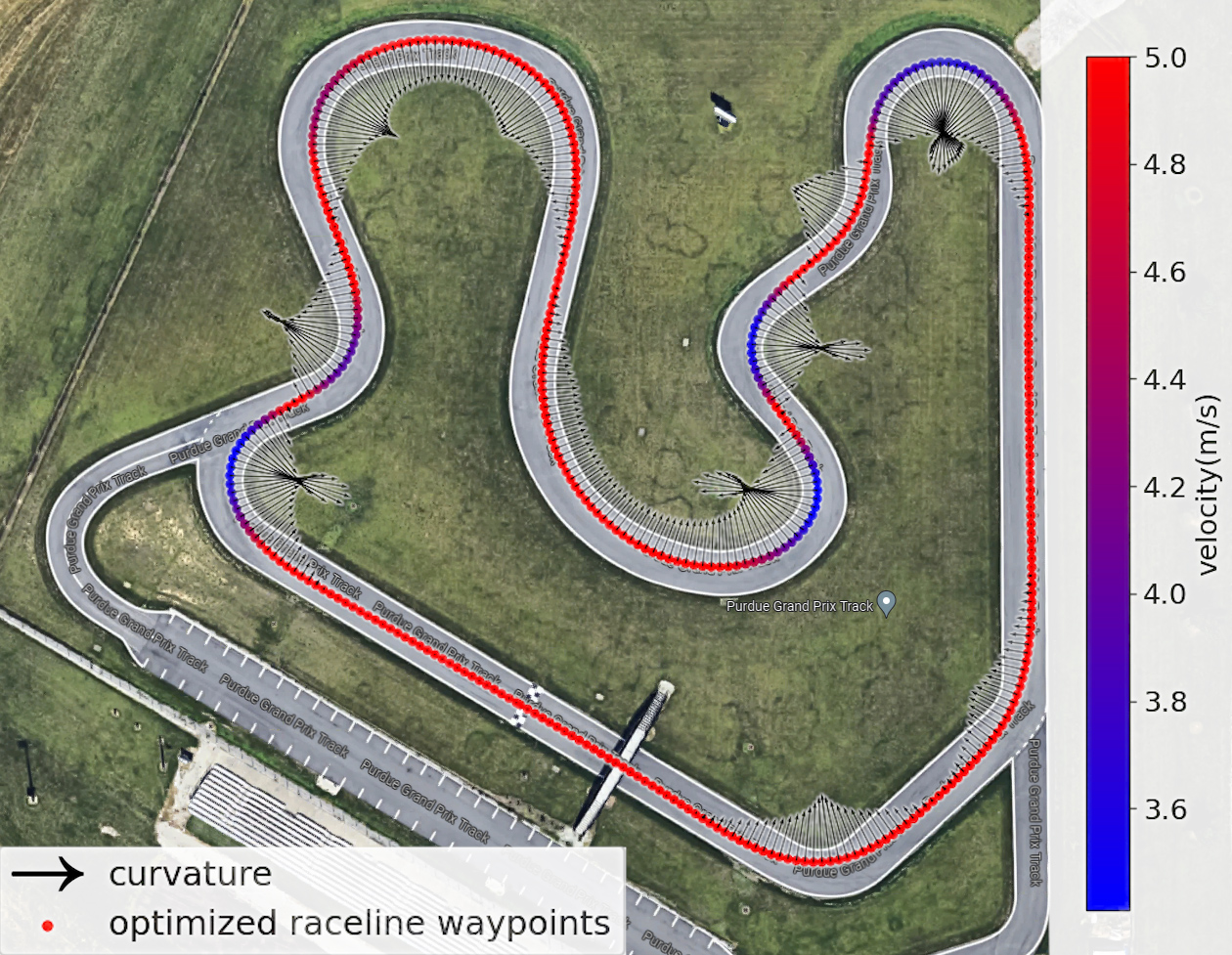} 
    \caption{Waypoints collection and raceline optimization at Purdue Grand Prix racing track, which spans a distance of 434 meters.}
    \label{fig:raceline}
    \vspace{-0.5cm}
\end{figure}

\subsection{Adaptive Pure Pursuit Controller}

To track the generated raceline, we implemented an adaptive pure-pursuit controller based on the geometric bicycle model \cite{sukhil2021adaptive}. Initially, a lookahead point is chosen on the raceline, situated at a fixed lookahead distance \(L\) from the vehicle. \(L\) is adaptively interpolated between a minimum \(L_{min}=2 \text{m}\) and a maximum \(L_{max}=5 \text{m}\), proportionally scaled to the vehicle's current velocity \(v\) and regulated by the maximum velocity \(v_{max} = 5 \text{m/s}\):
\begin{equation}
L = L_{min} + \frac{v}{v_{max}} (L_{max} - L_{min}).
\end{equation}

The lookahead point comprises both the desired velocity and position. Intuitively, for the vehicle to trace the arc from its current position to the lookahead point, the steering angle should be proportional to the arc curvature \(\gamma\). Utilizing geometric relationships, we deduced the radius \(r\) of the arc and subsequently determine \(\gamma\):
\begin{equation}
    \gamma = \frac{1}{r} = \frac{2|y|}{L^2},
\end{equation}
where \(|y|\) is the lateral distance from the vehicle to the lookahead point. To actuate the steering angle and enhance stability, we utilized a Proportional-Derivative (PD) controller that modulates the steering angle \(\delta_t\) according to \(\gamma\):
\begin{equation}
\delta_t = K_p \gamma_t + K_d\frac{d\gamma_t}{dt},
\end{equation}
where \(\gamma\) is treated as the cross-track error term \cite{snider2009automatic}, reflecting the lateral deviation. In practice, \(K_p = 2.0\), \(K_d = 1.0\). 

\subsection{Boundary Detection}

We devised a vision-based algorithm for detecting race track boundaries for the reactive component of the AKS competition, where pre-mapping was not permitted. The algorithm relies on grass detection surrounding the race track, employing classical computer vision techniques with OpenCV.

To identify grass regions, the input RGB camera image is blurred using a Gaussian filter to eliminate unwanted noise. Its blue and green channels are then extracted and normalized in grayscale, which grants higher intensities to green pixels than to pixels of other colors. Green pixels \(G(x, y)\) are identified by the green \(g\) and the blue \(b\) channel with a threshold \(\tau\), where \(\tau\) can be affected by many factors such as the environment and the lighting condition:
\begin{equation}
G(x, y) = \begin{cases} 
                1, & \text{if } 0.6 \cdot g - b \ge \tau \\
                0, & \text{if } 0.6 \cdot g - b < \tau
            \end{cases}.
\end{equation}
The resultant binary image \(G\) represents a mask for grass regions. This mask is then processed with open and then close morphology operations to remove small noise.

Next, we conducted a bird's-eye view (BEV) projection that converts an image from a front view to a top-down view. A transformation matrix is determined offline by mapping four points in the image to their respective BEV coordinates using OpenCV's getPerspectiveTransform function (Fig \ref{fig:grass}).

The final step is to convert the grass BEV into a 2D depth format. The depth data is denoted by a vector \(s \in \mathbb{R}^d\), where each \(s_i\) is a distance measurement from the go-kart to an object. Correspondingly, a vector \(a \in \mathbb{R}^d\) captures the angles associated with \(s_i\). The range of detection is \([-\pi / 2, \pi / 2]\), indicating a 180-degree field of view ahead of the vehicle sampled at \(0.5^\circ\) resolution. The zero angle is aligned with the vehicle's heading while angles are measured counter-clockwise. 

\subsection{Follow-the-Gap}

After acquiring depth data from boundary detection, we employed the FTG method to identify the largest gap that meets the required safety distance from the vehicle and navigate toward it. First, we defined a gap \(g\) as a continuous subsequence \([s_i, s_j]\) where \(i\) and \(j\) are the starting and ending indices respectively, such that:
\begin{equation}
\forall k \in [i, j], \ s_k \geq \epsilon,
\end{equation}
where \(\epsilon = 2.5 \text{m}\) is a safety distance threshold that determines the minimum allowable distance for a gap. We chose the largest gap as the optimal one, which starts at index \(i_{opt}\) and ends at index \(j_{opt}\). Then, we chose the midpoint of the optimal gap as the goal at index \(k_{mid}\) to reduce unnecessary oscillation:
\begin{equation}
k_{mid} = \frac{i_{opt}+j_{opt}}{2}.
\end{equation}
Since the zero angle is parallel to the vehicle's heading, we calculated the steering angle \(\delta\) from the angle vector \(a\):
\begin{equation}
\delta = a_{k_{mid}}.
\end{equation}

Thereafter, the desired velocity \(v\) is interpolated between a minimum \(v_{min}=2 \text{m/s}\) and a maximum \(v_{max}=5 \text{m/s}\), proportionally scaled to the vehicle's current steering angle \(\delta\), and regulated by the maximum allowable steering angle \(\delta_{max} = 1.0\text{rad}\):
\begin{equation}
v = v_{min} + \frac{\delta}{\delta_{max}} (v_{max} - v_{min}).
\end{equation}

\begin{figure}[!t]
    \includegraphics[width=\columnwidth]{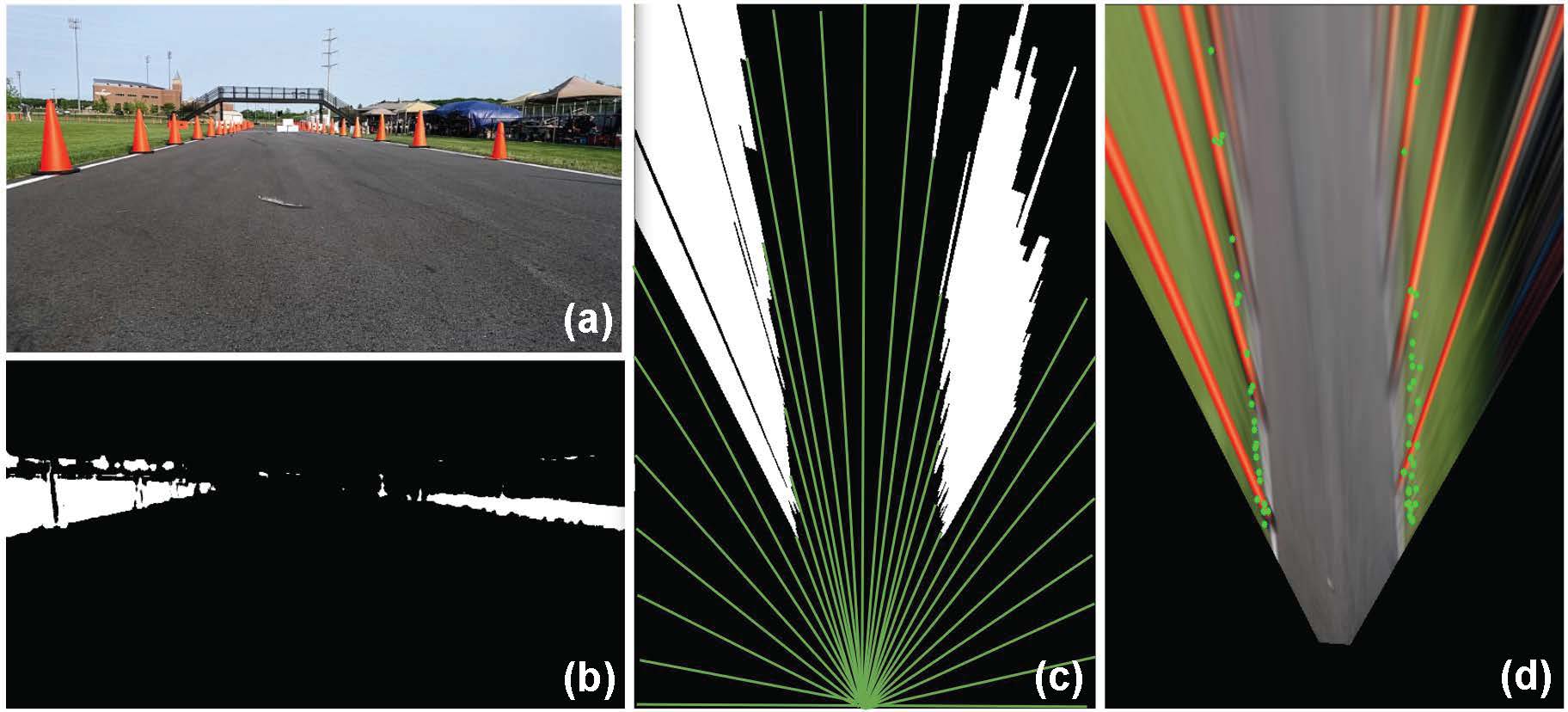} 
    \caption{Grass boundary detection. (a) Raw camera input. (b) Filtered grass mask. (c) The BEV of the grass mask. Green lines indicate the angles for searching grass distances. (d) The converted depth data is plotted as green dots and overlaid onto the BEV image of the camera input.}
    \label{fig:grass}
    \vspace{-0.5cm}
\end{figure}

\section{Conclusion}

In this paper, we introduced an open-source design for an electric go-kart platform enabling advanced research and development in autonomous driving systems. The design's modular mechatronic systems seamlessly support different driving modes. Additionally, we have implemented an adaptable sensor stack to execute tasks such as perception, localization, planning, and control. Our experimentation has showcased the go-kart's versatility, demonstrating its proficiency in the autonomous mode while running the pure pursuit and follow-the-gap algorithms. This innovative design effectively bridges the gap between reduced-scale cars and full-scale vehicles, enabling both widespread accessibility with high performance. It consequently provides immense value to universities and research institutions, fostering collaboration towards the open development and validation of autonomous vehicles.

Future work will focus on the continuous improvement of the mechatronic, sensing, and software systems. We plan to leverage the platform's different driving modes and explore human-machine interactions, such as the imitation learning algorithm \cite{f110_IL}, which involves dynamic cooperative control between the driver and the vehicle.

\section*{Acknowledgements}

The authors wish to express their gratitude to the Autoware Foundation for providing financial support for the project. 
Our thanks are also extended to Dr. Jack Silberman and his team from the University of California, San Diego for their contributions to the autonomous go-kart community. 
Lastly, we would like to thank Andrew Goeden for his efforts in organizing the 2023 AKS Purdue Grand Prix, providing an outstanding and enriching experience. 

\bibliography{main.bib}
\end{document}